\documentclass[11pt]{article}
\usepackage[hyperref]{ccl2021-en}
\usepackage{times}
\usepackage{url}
\usepackage{latexsym}
\usepackage{fancyhdr}
\usepackage{graphicx}
\usepackage{amsmath}

\title{Learn to Focus: Hierarchical Dynamic Copy Network for Dialogue State Tracking}

\author{Linhao Zhang,   Houfeng Wang\\ 
MOE Key Lab of Computational Linguistics, Peking University, Beijing, 100871, China\\ 
\{zhanglinhao, wanghf\}@pku.edu.cn\\
}

\date{}

\begin{document}
\maketitle

\begin{abstract}
Recently, researchers have explored using the encoder-decoder framework to tackle 
dialogue state tracking (DST), which is a key component of task-oriented 
dialogue systems. However, they regard a multi-turn dialogue as a flat sequence, 
failing to focus on useful information when the sequence is long. In this paper, we propose a Hierarchical Dynamic Copy Network (HDCN) to facilitate focusing on the most informative turn, making it easier to extract slot values from the dialogue context. Based on the encoder-decoder framework, we adopt a hierarchical copy approach that calculates two levels of attention at the word- and turn-level, which are then re-normalized to obtain the final copy distribution. A focus loss term is employed to encourage the model to assign the highest
turn-level attention weight to the most informative turn. Experimental results show that our model achieves 46.76\% joint accuracy on the MultiWOZ 2.1 dataset.
\end{abstract}

\section{Introduction}
Much attention has recently been paid to dialogue state tracking (DST), which aims to extract 
user goals at each turn of a dialogue and represents them using slot-value pairs. These pairs are later used by the policy 
learning module to decide the next system action.

With the development of deep learning, neural network-based methods have been widely applied to DST. Many previous studies \cite{Mrksic2016NeuralBT,Zhong2018GlobalLocallySD,Balaraman2019ScalableND} decompose DST into a series of classification problems. They first score all possible slot-values pairs and then 
choose the value with the highest probability as the predicted value 
for that slot. 

Some more recent work proposes generation-based DST \cite{Wu2019TransferableMS}.
Based on the encoder-decoder framework \cite{Sutskever2014}, it directly generates value for each slot.
 It is further enhanced 
with a soft copy mechanism \cite{see2017get}, which enables 
it to generate values using text from the input source.

\begin{figure}[t!]
\centering
\includegraphics[scale=0.4]{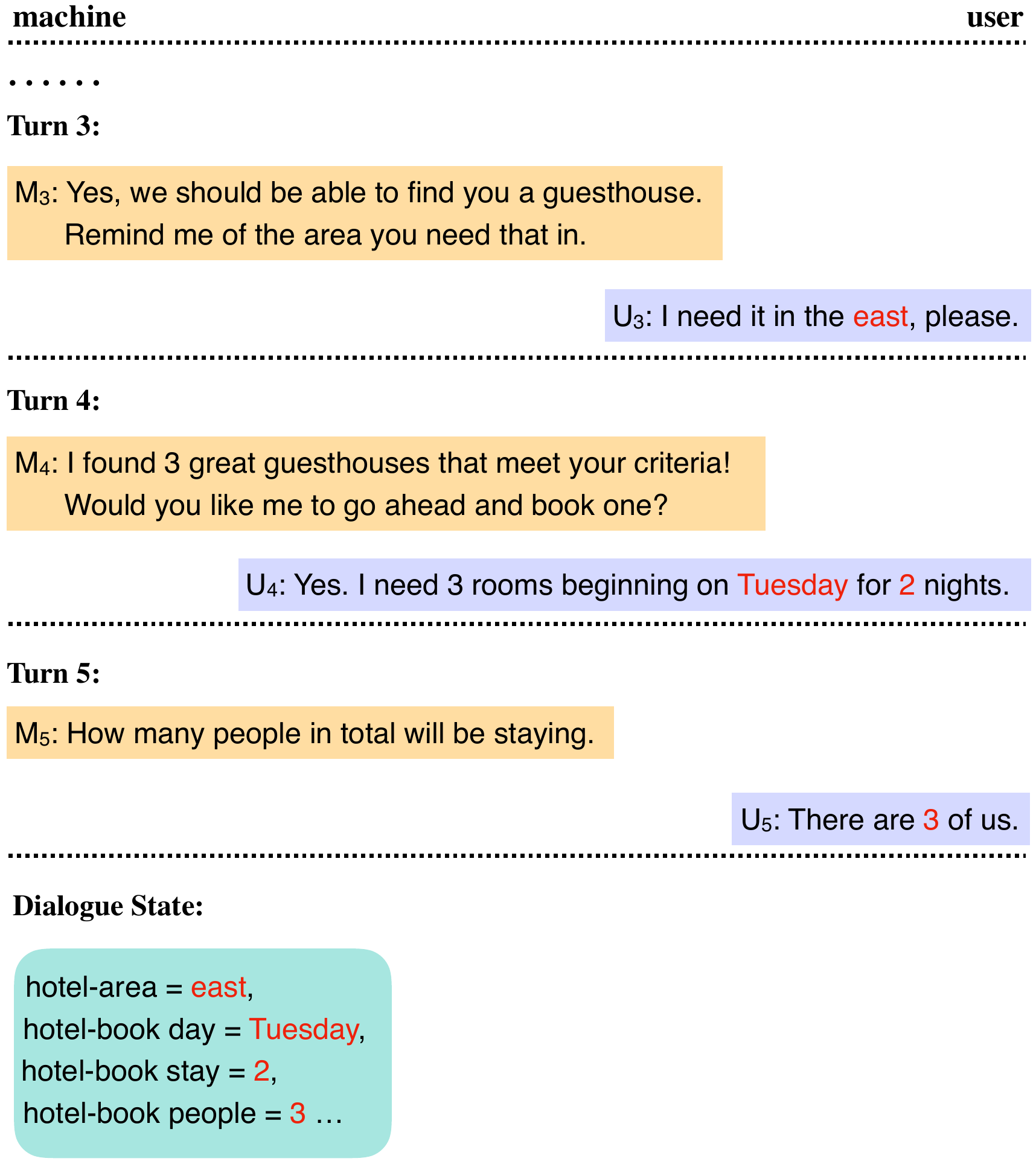}
\caption{A DST example from MultiWOZ 2.1 dataset. Dialogue state is composed of slot-value pairs. Note that throughout this paper, we use the term \emph{slot} to refer to the 
concatenation of \emph{domain} (e.g., hotel) and 
\emph{slot} (e.g., area), following the convention of MultiWOZ 2.1 dataset.}
\label{figure-example}
\end{figure}

Despite its remarkable success, \cite{Wu2019TransferableMS} simply concatenates different dialogue turns and 
feeds them as a flat sequence to the model. 
For example, it would organize the dialogue in Figure \ref{figure-example} in
the form of $M_1U_1M_2U_2M_3U_3...$.
Obviously, when the dialogue contains multiple turns, the sequence would become long, and copying the correct slot values becomes 
especially difficult.

To this end, we propose a Hierarchical Dynamic Copy Network (HDCN) to facilitate focusing on the most informative 
turns, making it easier to copy slot values from the dialogue context. We define the \emph{information turn} of a certain slot as 
the turn where its corresponding value appears, or can be inferred. For 
example, in Figure \ref{figure-example}, the \emph{information turn} 
for slot \emph{hotel-area} is turn 3, where its value \emph{east} appears. Obviously, 
to generate value for a certain slot, we should focus more on its \emph{information turn}. 

To achieve this, we use two levels of 
attention applied to the word- and turn-level, which are 
then re-normalized and result in the final copy distribution.
We further add a focus loss term to encourage our model to assign higher turn-level 
attention weight to the \emph{information turn}. 
Notably, our turn representation is calculated dynamically during decoding 
using the word-level attention. This dynamic calculation leads to better 
turn representations for two reasons: 1) it utilizes all the hidden states 
instead of only the last one, and 2) it enables our model 
to form different turn representations for different slots. As shown in Figure \ref{figure-example}, when calculating word-level attention over turn 4, the slot \emph{hotel-book day} 
would attend more to the word \emph{Tuesday}, and the slot \emph{hotel-book stay} would assign more attention to the word \emph{2}. In this way, the resulting 
turn-representations for these two slots also become different.

We conduct experiments on the MultiWOZ 2.1 dataset. Results show that HDCN brings over 1\% (absolute) 
improvement in terms of joint accuracy, with better interpretability. Note that our goal is not to 
beat the state-of-the-art systems such as \cite{kim2020somdst,hosseini2020simple}, as they are often based on pretrained language model (e.g., BERT).  Rather, we aim to offer a lightweight improvement over the best system in the pre-BERT era \cite{Wu2019TransferableMS}. We believe although BERT-based models can achieve high accuracy, they inevitably cost more inference time, making the DST component not practical in real-word dialog systems.



\section{Model}
\subsection{Task Definition}
Given a set of machine and user utterance pairs in \emph{T}
turns of dialogue $X = \{(M_1,U_1),...,(M_T,U_T)\}$, we aim 
to generate dialogue states $B = \{(S_1,V_1),...,(S_M,V_M)\}$ containing a series of slot-value pairs $(S_m,V_m)$, where $M$ is the 
number of all the possible slots. 
In the following subsection, we first give a brief description of generation-based DST and  introduce our work in detail.

\subsection{Generation-Based DST}
\label{subsection-baselines}
\cite{Wu2019TransferableMS} first proposes to regard DST as a generation problem. Their proposed model, \emph{TRADE}, achieves the best performance in the pre-BERT era. It follows the encoder-decoder paradigm \cite{Sutskever2014}, and adopts a soft-copy mechanism \cite{see2017get} to copy slot values from the input source. 

At encoding time, it uses a bi-directional gated recurrent units (GRU) \cite{cho2014learning} to encode the dialogue history, which is organized as a flat sequence $M_1U_1M_2U_2...M_TU_T$. 

To generate value $V_m$ for slot $S_m$, it supplies the \emph{summed embedding} of $S_m$ as the first input to the decoder\footnote{More specifically, the \emph{summed embedding} of the $m_{th}$ domain and slot. For example, if $S_m$ is \emph{hotel-area}, then the summed embedding of $S_m$ is \emph{Emb}(hotel) + \emph{Emb}(area)}, and decodes $M$ times independently, where $M$ is the number of all the possible slots.

At decoding step $k$ , the decoder returns a hidden 
state $h_{k}$, which is later mapped into the vocabulary space 
$P_{k}^{vocab}$ using embedding matrix $E$. Formally,

\begin{equation}
P_{k}^{vocab}=softmax\left(E \cdot(h_k)^{\top}\right)
\label{equation-vocab}
\end{equation}

At the same time, $h_k$ is used to
calculate attention weights $a_i^k$ over the encoded dialogue history. $a_{i}^k$ and $P_k^{vocab}$ are then weighted and summed to obtain the final word distribution \cite{see2017get}:

\begin{equation}
P_k(w)=p_{\mathrm{gen}} P_k^{\mathrm{vocab}}(w)+\left(1-p_{\mathrm{gen}}\right) \sum_{i  : w_{i}=w} a_{i} ^{k}
\end{equation}
\begin{equation}
p_\mathrm{gen}=\sigma\left(W_{p} \cdot\left[h_{k} ; e_{k} ; c_{k}\right]\right)
\label{equation-gen}
\end{equation}
where $W_p$ is a trainable matrix, $e_k$ is the input word embedding to the decoder at time step $k$ and $c_k$ is the context vector \cite{bahdanau2014neural}. Intuitively, $p_{gen}$ can be seen as a soft switch to choose between generating a word from the vocabulary, and copying a word from the 
dialogue history.

\emph{TRADE} also contains a slot gate, which is a simple
three-way classifier that maps the context vector to a probability distribution over 
\emph{ptr}, \emph{none}, and \emph{dontcare} classes. 
For more details, please refer to the original paper \cite{Wu2019TransferableMS}.





\subsection{HDCN}
\label{subsection-HDCN}
\begin{figure*}[t!]
\centering
\includegraphics[scale=0.3]{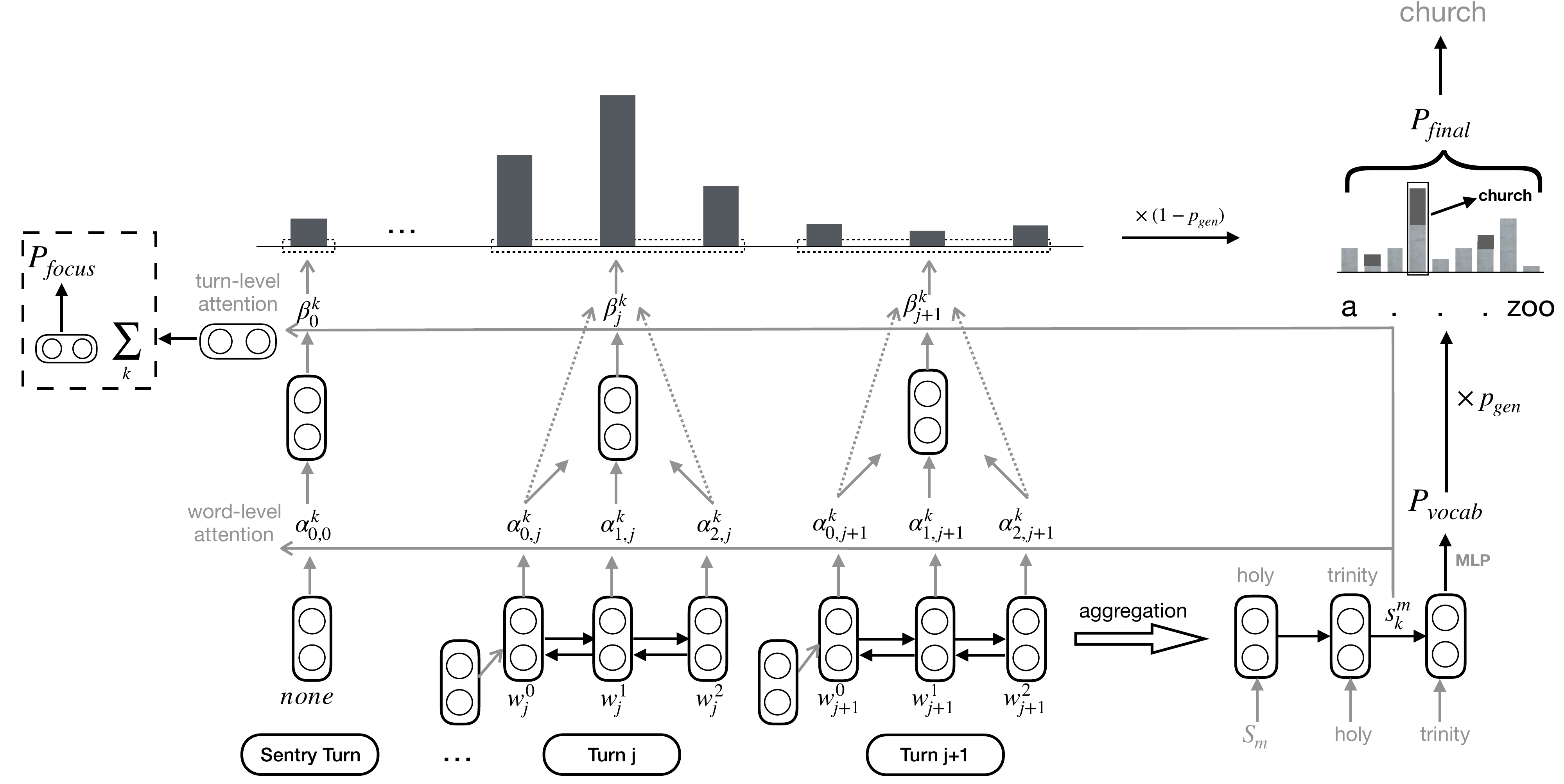}
\caption{The architecture of HDCN. It follows the encoder-decoder 
framework, with the dialogue as input to the encoder and the target slot $S_m$ as the first input to the decoder. It has two levels of attention applied at the word- and turn-level, which are then re-normalized and result in the final copy distribution. The model is further enhanced with a focus mechanism to encourage focusing on the \emph{information turn} of the target slot. We pad a sentry turn at the beginning of the dialogue, which acts as the \emph{information turn} of slots never mentioned in the dialogue.}
\label{figure-model}
\end{figure*}

Although \emph{TRADE}  takes an important first step towards generation-based DST, it 
simply concatenates different dialogue turns and regards the multi-turn dialogue as a 
flat sequence. Correspondingly, copying the correct slot values becomes especially hard when the sequence is long.
In this paper, we propose the hierarchical dynamic copy network (HDCN) to address this 
limitation. The overall structure of our model is shown in Figure \ref{figure-model}.

We first encode each turn with a bidirectional LSTM (BiLSTM) encoder and summarize the information in the forward and backward BiLSTM hidden 
states, namely $h_{i}^j=[\overrightarrow{h_{i}^j} ; \overleftarrow{h_{i}^j}]$. 
During encoding, we compare two initialization 
strategies for $h_0^j$:

\begin{enumerate}
\item  \emph{zero initialization}: As in \cite{Serban2015BuildingED,Serban2016AHL}, we initialize $h_0^j$ with zero vector.
\item \emph{last initialization}: $h_0^j$ is initialized as the last hidden state of previous turn $h^{j-1}$.
\end{enumerate}

After encoding, we aggregate all the encoded information 
into a fix-length vector using max-pooling, with which we initialize our LSTM decoder. 
As in \emph{TRADE}, our model decodes $M$ times independently to generate the dialogue state.
For the slot $S_m$ at time step $k$:

\begin{equation}
s_k^m = LSTM(e_k^m,s_{k-1}^m)
\label{equation-decoder}
\end{equation}
where $e_k^m$ is the word embedding and $s_k^m$ is the decoder hidden state.
Following \emph{TRADE}, we also supply the \emph{summed embedding} of $S_m$
as the first input, $e_0^m$, to the decoder.

After obtaining $s_k^m$,  we adopt a hierarchical dynamic copy approach to obtain copy distribution. 
Specifically, we first calculate word-level attention distribution $\alpha_{ij}^k$
over the dialogue history \cite{luong2015effective}:

\begin{equation}
\alpha_{ij}^{k} = softmax(s_k^mW_wh_i^j) \\
\label{equation-word-level}
\end{equation}
where $W_w$ is trainable parameters.

Note that $\alpha_{ij}^{k}$ is dependent on the target slot $m$ (we skip $m$ in the notations $\alpha_{ij}^k$ for simplicity.).
Intuitively, the decoder hidden state $s_k^m$ can be seen as a high 
level representation of a fixed query:
\emph{what is the informative words for the $m_{th}$ slot}? 

We then aggregate the representations of those informative words to form turn representations:

\begin{equation}
g_{j}^{k} = \sum_{i}\alpha_{ij}^{k}h_i^j \\
\end{equation}

Traditional hierarchical LSTM methods regard the last hidden state of the word-level 
LSTM as the turn representation. By contrast, our turn representation considers all the 
hidden states $h_i^j$. More importantly, this dynamic calculation 
procedure enables our model to form different turn representations $g_j^{k}$ for 
different target slots, leading to more flexible and robust representations. 
For example, in Figure \ref{figure-example}, when calculating $\alpha_{ij}^k$ over 
turn 4, the slot \emph{hotel-book day} would attend more to the 
word \emph{Tuesday}, and the slot \emph{hotel-book stay} would attend more to the word \emph{2}.
In this way, the turn representations $g_j^k$ for these two slots also differ.

Similarly, to identify informative turns for the $m_{th}$ slot, we calculate 
turn-level attention weights using $s_k^m$. 

\begin{equation}
\beta_{j}^{k} = softmax(s_k^mW_tg_j^k) 
\label{equation-turn-level}
\end{equation}
where $W_t$ is model parameters.
Intuitively, $\beta_j^{k}$ can be seen as an indicator of \emph{how informative turn 
j is for the $m_{th}$ slot} (similarly, we skip $m$ in the notation for simplicity). 

Then, we re-normalize the two levels of attention to  obtain the final copy 
distribution. Formally,

\begin{equation}
\gamma_{i j}^{k}=\beta_{j}^{k} * \alpha_{i j}^{k}
\label{equation-normalization}
\end{equation}

In this way, words from more informative turns get rewarded, and words from 
less informative turns are discouraged. A more detailed analysis can be found in 
Subsection \ref{subsection-attention}.

Finally, $\gamma_{ij}^{k}$ and the vocabulary 
distribution $P^{vocab}_k$ are weighted and summed to obtain the 
final word distribution:

\begin{equation}
P_k(w)=p_{\mathrm{gen}} P^{\mathrm{vocab}}_k(w)+\left(1-p_{\mathrm{gen}}\right) \sum_{i j : w_{i j}=w} \gamma_{i j}^{k}
\end{equation}
where $P_k^{vocab}(w)$ and $p_{gen}$ are calculated similar to Equation \ref{equation-vocab} and \ref{equation-gen}, respectively.

\subsection {Turn Focus Mechanism}
\label{subsection-focus}
We define the \emph{information turn} for a certain slot as the turn where its 
corresponding value appears, or can be inferred.  It is essential to 
assign the highest turn-level attention weight to the \emph{information turn} in order to capture key information. 

To achieve this, we sum and normalize all the turn-level 
attention weights after decoding, and then calculate the turn focus loss as 
the negative log probability of the
information turn.  Formally,

\begin{equation}
\beta_j = softmax(\sum_{k} \beta_j^{k})
\end{equation}

\begin{equation}
\mathcal{L}_{focus}^m=-\log (\hat{\beta_j})
\label{equation-tl}
\end{equation}
where $\hat{\beta_j}$ is the \emph{information turn} for the $m_{th}$ slot.

In this way, our model can learn to assign the highest turn-level 
attention weight to the \emph{information turn}. Combined with our hierarchical 
dynamic copy mechanism, words from the \emph{information turn} are more likely to be 
copied (see Equation \ref{equation-normalization}). 

Note that some slots are never mentioned in the dialogue, and they thus do not have 
corresponding \emph{information turns}. To handle these slots, we pad a sentry turn at 
the beginning of the dialogue (see Figure \ref{figure-model}). The sentry turn contains only one word \emph{none} and 
acts as the \emph{information turn} for these slots.

Besides, we observe that slot values that contain multiple words are mostly named entities mentioned in the dialogue, like \emph{holy trinity church} or 
\emph{london liverpool street}. Obviously, these words are almost 
always from the same turn. Based on this observation, we further 
propose two methods to better handle multi-words values:  \emph{HDCN-freeze} and 
\emph{HDCN-force}. 

As the name suggests, \emph{HDCN-freeze} freezes turn-level attention 
after the first decoding step. That is, we modify Equation \ref{equation-turn-level} into the following:

\begin{equation}
\beta_j^{k}=\left\{\begin{array}{ll}{softmax(s_k^mW_tg_j^k) } & {\text { if k $=$ 0}}  \\ {\beta_j^0} & {\text{ if k $>$ 0}}\end{array}\right.
\end{equation}

However, we found that this approach is too rigid.
Inspired by the coverage mechanism \cite{see2017get}, we further propose \emph{HDCN-cover}, a more 
flexible approach. The basic idea is to \emph{encourage} rather than \emph{
force} our model to focus on the same turn during decoding. 
Specifically, we maintain a turn coverage vector $c^k$ during decoding, which is the sum of 
turn-level attention distribution over all previous decoding steps:

\begin{equation}
c^k=\sum_{k=0}^{k-1} \beta^{k}
\end{equation}
$c^k$ is used as extra input to the calculation of turn-level attention, changing equation \ref{equation-turn-level} to:

\begin{equation}
\beta_{j}^{k} = softmax(s_kW_kg_j + w_cc_j^k) 
\end{equation}
where $w_c$ is a learnable parameter randomly initialized. 
During training, we find that $w_c$ steadily increases, suggesting 
that turns receiving higher attention in previous decoding steps indeed 
get rewarded. 

Note that our coverage vector serves the exact opposite purpose of 
the original coverage mechanism \cite{see2017get}. Coverage mechanism 
was first proposed for repetition reduction by \emph{discouraging} the 
model from attending to the same place. By contrast, we here want to
\emph{encourage} the model to attend to the same turn during decoding.

\subsection{Task Learning}
During training, the cross-entropy loss for the $m_{th}$ slot is calculated as:
\begin{equation}
\mathcal{L}_{ce}^m=-\frac{1}{K}\sum_{k=1}^K \log P\left(\hat{w}_{k}^m\right)
\end{equation}
where \(\hat{w}_{k}^m\) is the ground truth at time step \(k\) and $K$ is the total decoding steps.

Combined with Equation \ref{equation-tl}, the final loss function is defined as:

\begin{equation}
\mathcal{L}=\sum_{m = 1}^M(\mathcal{L}_{c e}^m+\xi \mathcal{L}_{focus}^m)
\label{equation-loss}
\end{equation}

where $\xi$ is the focus ratio and is used to adjust the relative importance of each term.  $M$ is the number of all the possible slots.



\section {Experiments}
\subsection{Dataset}
Following previous work \cite{Wu2019TransferableMS,Ren2019ScalableAA,Gao2019DialogST}, 
we train and evaluate our model with MultiWOZ \cite{Budzianowski2018MultiWOZA} 
dataset. It is currently the largest 
multi-domain conversational corpus spanning over seven 
domains, containing 8438 multi-turn dialogues. The training set, validation set and 
test set contain 8,438, 1,000 and 1,000 dialogues, respectively.

The original MultiWOZ dataset were detected to contain substantial 
errors in the state annotation. These errors were later fixed in
MultiWOZ 2.1 dataset \cite{Eric2019MultiWOZ2M}. The correction process results in 
changes to over 32\% of state annotations across 40\% of the dialogue turns.
In this paper, we conducted experiments with the corrected version dataset, MultiWOZ 2.1.

Note that MultiWOZ dataset does not contain \emph{information turn} labels, yet it can 
be easily inferred using rule-based method. Specifically, if the 
dialogue state changes from  $B_{t-1}$  to $B_{t}$ after turn $t$, then $t$ is the 
\emph{information} turn for those slots whose values differ in $B_{t-1}$ and $B_t$. 
The \emph{information turn} labels are only used during training.

\subsection{Training Procedure}
Our model was implemented in PyTorch and trained on a single NVIDIA 1080 Ti GPU. 
The LSTMs \cite{hochreiter1997long}  and word embeddings are 
400 dimensional. 
Following previous work \cite{Zhong2018GlobalLocallySD,Wu2019TransferableMS}, we initialized the embeddings by concatenating Glove embeddings \cite{Pennington2014GloveGV}
and character embeddings \cite{Hashimoto2016AJM}.
Dropout \cite{hinton2012improving} is used with dropout rate set to 
0.5. The number of layers of LSTM encoder/decoder is set to 2. The 
batch size is set to 16. $\xi$ in Equation \ref{equation-loss} is set to 0.1.

We use Adam \cite{kingma2014adam} with
learning rate = $10^{-3}$, $ \beta_{1}=0.9, \beta_{2}=0.98$ and $\epsilon=10^{-9}$. 
We choose our model according to the results on the development set. 
The reported results are the average of five test set results.

\subsection{Systems for Comparison}
In this paper, we compare our model against the following baselines:

\textbf{FJST} \cite{Eric2019MultiWOZ2M} is a classification-based model. It uses bidirectional LSTM network to encode the full dialogue history.

\textbf{HJST}  \cite{Eric2019MultiWOZ2M} differs from FJST in that it 
encodes the dialogue using hierarchical LSTM \cite{Serban2015BuildingED}. 



\textbf{DST Reader} \cite{Gao2019DialogST} first proposes to regard the problem of DST as a reading comprehension problem. 
It uses a simple attention-based neural network to point to the slot values within the dialogue.

\textbf{HyST}  \cite{Goel2019HySTAH} combines a hierarchical encoder fixed vocabulary system with an 
open vocabulary n-gram copy-based system     


\textbf{TRADE}  \cite{Wu2019TransferableMS} is the first generation-based DST model and  achieves the state-of-the-art performance in the pre-BERT era.  A detailed description is in   Subsection \ref{subsection-baselines}  


\textbf{Hier-LSTM based} is our baseline to examine the effectiveness of our hierarchical dynamic copy mechanism. It is based on the hierarchical LSTM encoder, and the decoding process is the same as \emph{TRADE}. For a fair comparison, we also implement copy and 
focus mechanism to this baseline. 

\textbf{HDCN} is our proposed model, and we compare four variants of HDCN in this paper. The difference between zero- and last-initialization is  elaborated in Subsection \ref{subsection-HDCN} and the difference between \emph{HDCN-freeze} and \emph{HDCN-cover} is detailed in Subsection \ref{subsection-focus}.

\subsection{Metrics}
\textbf{Joint accuracy -}
Joint accuracy is the most widely used evaluation metrics for
DST \cite{Zhong2018GlobalLocallySD,Wu2019TransferableMS,Ren2019ScalableAA}.
It compares the predicted dialogue state to the 
ground truth at each dialogue turn, and the output is considered 
correct if and only if all the predicted values exactly match the 
ground truth values. All the baseline results are taken from the official paper of MultiWOZ 2.1 dataset \cite{Eric2019MultiWOZ2M}.

\textbf{Focus accuracy -}
In this work, we also consider focus accuracy to examine our 
claim that the focus mechanism can encourage the model to focus on 
\emph{information turns}. For each slot, focus accuracy compares 
its summed turn-level attention weights $\sum_{k=0}^K\beta^k_j$ with its 
\emph{information turn}. The result is considered correct if and only if the 
\emph{information turn} receives the highest attention weight.

\begin{table}[t!]
\begin{center}
\scalebox{1}{
\begin{tabular}{|l|c|}
\hline
\textbf{Models}   & \textbf{Joint Acc.} \\ \hline 
DST Reader \cite{Gao2019DialogST}  & 36.40 \\
HyST    \cite{Goel2019HySTAH}      & 38.10 \\  
FJST    \cite{Eric2019MultiWOZ2M}     & 38.00 \\   
HJST     \cite{Eric2019MultiWOZ2M}     & 35.55 \\   
TRADE     \cite{Wu2019TransferableMS}    & 45.60 \\ \hline \hline
Hier-LSTM based (baseline)  & 39.39 \\ 
HDCN-freeze, zero initialized  & 45.48 \\  
HDCN-freeze, last initialized  & 46.11 \\  
HDCN-cover, zero initialized   & 45.62 \\  
HDCN-cover, last initialized   & \textbf{46.76} \\  \hline
\end{tabular}
}
\end{center}
\caption{\label{table-accuracy} Joint accuracy on MultiWOZ test set.} 
\label{table-results}
\end{table}

\begin{table}[t!]
\begin{center}
\scalebox{1}{
\begin{tabular}{|l|c|c|}
\hline
\textbf{Settings} &\textbf{Focus Acc.}  & \textbf{Joint Acc.}\\ \hline 
$\xi=0$    &94.89   & 50.23  \\  
$\xi=0.01$ & 95.70 & 51.06\\
$\xi=0.05$ & 96.17 & 51.18\\
$\xi=0.1$  &96.31 & \textbf{51.32}  \\
$\xi=0.2$  &96.47  & 50.80 \\
$\xi=1$  &\textbf{96.63} & 48.87 \\ \hline

\end{tabular}
}
\end{center}
\caption{\label{table-ablation} Development set results to examine the effectiveness of our focus mechanism. } 
\label{table-focus}
\end{table}

\section{Results}
\subsection {Overall Performance}
\label{sect-overall}
The joint accuracy results on the MultiWOZ dataset are shown 
in Table \ref{table-results}. 

Our first observation is that the \emph{Hier-LSTM based} baseline gives
poor performance, with joint accuracy lower than the 
flat-structured \emph{TRADE}. Besides, we can see that \emph{HJST} also performs 
worse than \emph{FJST}, which further proves that simply using two-levels of 
LSTM to model multi-turn dialogue is not suited for DST. 
We suppose this is because the last hidden state of word-level LSTM is not sufficient 
to capture all the key information within the turn.
By contrast, our proposed model uses the attention mechanism to 
dynamically calculate turn representations and selectively uses all the word 
representations within the turn. More importantly, our model allows for different turn 
representations for different target slots. 

Besides, we can see that \emph{HDCN-cover} generally performs better than 
\emph{HDNC-freeze}. Although they both aim for consistent focusing on the same turn 
during decoding, the former approach allows for more flexibility. 
Another observation is that the \emph{last-initialization} approach performs notably 
better than \emph{zero-initialization}, with about 1\% absolute improvement. 
This is because the former can effectively utilize history 
information when encoding each turn, while the latter encodes each turn independently and discards the history information entirely.

\begin{figure*}[t!]
\centering
\includegraphics[scale=0.4]{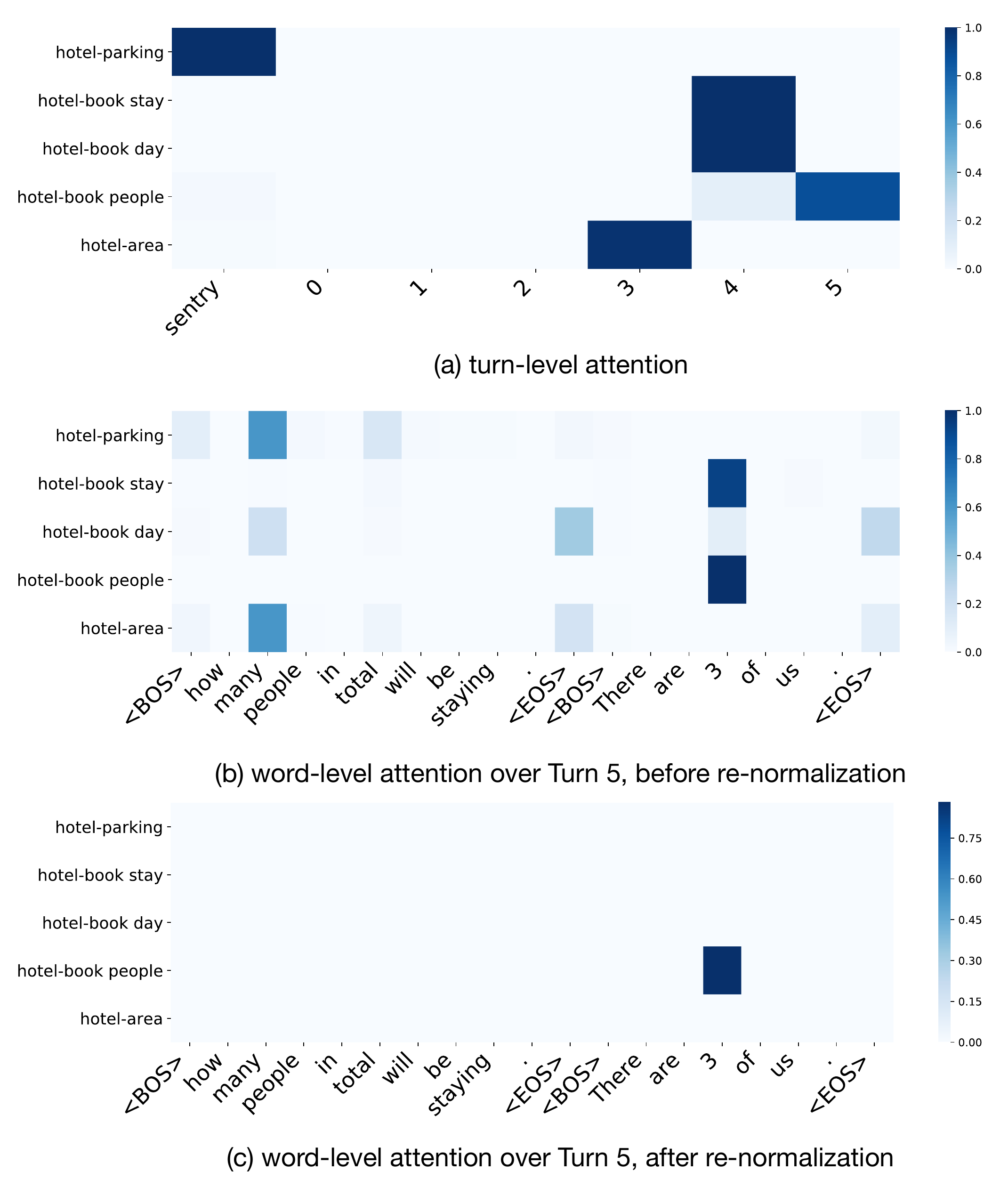}
\caption{The turn- and word-level attention weights.} 
\label{figure-case}
\end{figure*}
\subsection {Focus Mechanism}
One of the main claims of this paper is that we can improve 
DST performance by encouraging our model to focus on the \emph{information turn}. 
To further examine this claim, we conduct experiments with different
focus ratio, i.e., $\xi$ in  Equation \ref{equation-loss} on the development set. 
The results are shown in Table \ref{table-ablation}.

Firstly, we can see that the focus accuracy indeed increases with   
the focus ratio. This is understandable because a higher focus ratio means 
we attach greater importance to the focus loss term.

Besides, the results also buttress our claim that the focus mechanism can improve 
DST performance. We can see that even with 0.01 focus ratio, the improvement is 
notable. Our model reaches the best performance when $\xi=0.1$ and the reported 
test set results are based on this setting. 

Furthermore, we note that the joint accuracy does not increase 
monotonously with the focus ratio. After reaching its peak value, the 
joint accuracy slightly decreases. We suppose this is because 
focusing too much on a single turn may lead to the neglect of 
contextual information. Besides, from Equation \ref{equation-loss} we 
can see that excessively high focus ratio could hinder the decoding 
process.

\subsection {Attention Weights Analysis}
\label{subsection-attention}
We show in Figure \ref{figure-case} the turn- and word-level attention weights for 
the example in Figure \ref{figure-example}. 

As demonstrated in Subgraph (a),  all the five slots successfully focus on their corresponding \emph{information slots}.
For example, slot \emph{hotel-book day} assigns the highest attention weight
to turn 4, and slot \emph{hotel-book people} assigns the highest 
attention weight to turn 5. Besides, since the slot \emph{hotel-parking} is never 
mentioned in the dialogue, it assigns the highest attention weight to the sentry turn, 
which only contains one word \emph{none}.

Subgraph (b) gives the word-level attention over turn 5 before 
re-normalization with Equation \ref{equation-normalization}. 
A crucial observation is that both slot \emph{hotel-book stay} and \emph{hotel-book people} assign the highest attention weight to the word \emph{3}. 
This is because both slots tend to receive a number as value, and the number \emph{3}
can hence become puzzling. In fact, we find that 
\emph{TRADE}, which adopts flat-copy and does not involve the re-normalization process, wrongly assigns the value for slot \emph{slot-book stay} to be \emph{3}.

Subgraph (c) gives the word-level attention over turn 5 after re-normalization. We can 
see that since slot \emph{hotel-book stay} barely attends to turn 5 
(see Subgraph (a)), its attention on the word \emph{3} almost becomes 0 after 
re-normalization. 
By contrast, slot \emph{hotel-book people} still (rightfully) assigns
the highest attention to \emph{3}, because turn 5 is its \emph{information turn} and receives the highest turn-level attention.

From the above analysis, we can see why the proposed model improves over 
\emph{TRADE}. 
\emph{TRADE} organizes the dialogue as a long sequence. It is hence hard for it to 
attend to and copy the \emph{gold value}, as there can be many puzzling 
words along that sequence, like \emph{3} for slot \emph{hotel-book stay} in the above 
example. By contrast, our model adopts a hierarchical copy process and can assign the highest turn-level attention weight to the \emph{information turn}, thanks to the 
focus mechanism.



\section{Related Work}
\textbf{DST - } In recent years, neural network approaches have 
defined the state-of-the-art in DST research. Most previous works 
decomposed this task as a series of classification problems \cite{Zhong2018GlobalLocallySD,Mrksic2016NeuralBT,Ramadan2018LargeScaleMB,Balaraman2019ScalableND,Lee2019SUMBTSM}. 
They took each of the slot-value pairs as input for scoring, and 
output the value with the highest score as the predicted value for a 
slot. For such models, a predefined ontology is required, which lists 
all the possible slot-value pairs. However, such an ontology is not 
always available in practice. Besides, the number of all the possible 
slot-values pairs can be large, and enumerating all the possible slot-values pairs sometimes requires astronomical time and resources.
For example, in MultiWOZ dataset, there are over 4,500 possible slot 
values in total, meaning that such models have to perform over 4,500 
classifications to determine the current dialogue state.

To address these limitations, some recent works discarded the 
classification framework \cite{Xu2018AnEA,Wu2019TransferableMS,Balaraman2019ScalableND,Gao2019DialogST}.  \cite{Xu2018AnEA} is the first model that applies 
pointer networks \cite{Vinyals2015Pointer} to the single-domain DST problem, which generates both start and end 
pointers to perform index-based copying. \cite{Gao2019DialogST}  used a simple attention-based neural network to point to the slot values within the conversation. 
\emph{TRADE}  \cite{Wu2019TransferableMS} first proposed to regard 
this problem as a text generation problem. It encodes the dialogue 
history as a long sequence with a GRU and uses another GRU as the 
decoder to predict the value for each slot independently. \cite{Ren2019ScalableAA} proposed to sequentially decode domain, slot and 
value, hence remained a constant computational complexity, regardless 
of the number of predefined slots. 

\textbf{Hierarchical Attention Network -}  Our hierarchical dynamic copy mechanism 
draws inspiration from Hierarchical Attention Network (HAN) 
\cite{yang2016hierarchical}, which was first proposed for text classification. Similar 
to HAN, our approach also has two levels of attention mechanisms 
applied at the word- and sentence-level. However, our approach is notably different 
from HAN in two ways: one is that that when calculating attention scores (Equation \ref{equation-word-level} and \ref{equation-turn-level}), we use the decoder hidden state $s_t^m$ while HAN uses randomly 
initialized vector. In this way, our model can form different representations 
for different target slots. The other difference is that we  re-normalize  the two 
levels of attention to form the final copy distribution, while HAN only uses attention 
distribution to form sentence and document representations.

\textbf{Copy Mechanism} There are three common methods to perform copying, namely, index-based copy \cite{Vinyals2015Pointer}, hard-gated copy \cite{Gulcehre2016Pointing}and soft-gated copy \cite{see2017get}. \cite{Wu2019TransferableMS} 
first introduced soft-copy into DST area.  We extend their idea from flat copy 
to hierarchical dynamic copy.

\section{Conclusion}
In this work, we propose to tackle the DST problem with HDCN. Instead of viewing the 
dialogue as a flat sequence, we propose the hierarchical dynamic copy mechanism to facilitate focusing on the \emph{information turns}, making it easier to copy slot values from the dialogue context. Besides, we further propose two methods to better handle multi-words values. Extensive experiments show that our model improve over TRADE over 1\%. 


\bibliographystyle{ccl}
\bibliography{ccl}

\begin{thebibliography}{}

\bibitem[\protect\citename{Bahdanau \bgroup et al.\egroup
  }2014]{bahdanau2014neural}
Dzmitry Bahdanau, Kyunghyun Cho, and Yoshua Bengio.
\newblock 2014.
\newblock Neural machine translation by jointly learning to align and
  translate.
\newblock {\em arXiv preprint arXiv:1409.0473}.

\bibitem[\protect\citename{Balaraman and Magnini}2019]{Balaraman2019ScalableND}
Vevake Balaraman and Bernardo Magnini.
\newblock 2019.
\newblock Scalable neural dialogue state tracking.
\newblock {\em ArXiv}, abs/1910.09942.

\bibitem[\protect\citename{Budzianowski \bgroup et al.\egroup
  }2018]{Budzianowski2018MultiWOZA}
Pawel Budzianowski, Tsung-Hsien Wen, Bo-Hsiang Tseng, I{\~n}igo Casanueva,
  Stefan Ultes, Osman Ramadan, and Milica Gasic.
\newblock 2018.
\newblock Multiwoz - a large-scale multi-domain wizard-of-oz dataset for
  task-oriented dialogue modelling.
\newblock In {\em EMNLP}.

\bibitem[\protect\citename{Cho \bgroup et al.\egroup }2014]{cho2014learning}
Kyunghyun Cho, Bart Van~Merri{\"e}nboer, Caglar Gulcehre, Dzmitry Bahdanau,
  Fethi Bougares, Holger Schwenk, and Yoshua Bengio.
\newblock 2014.
\newblock Learning phrase representations using rnn encoder-decoder for
  statistical machine translation.
\newblock {\em arXiv preprint arXiv:1406.1078}.

\bibitem[\protect\citename{Eric \bgroup et al.\egroup
  }2019]{Eric2019MultiWOZ2M}
Mihail Eric, Rahul Goel, Shachi Paul, Abhishek Sethi, Sanchit Agarwal, Shuyang
  Gao, and Dilek~Z. Hakkani-T{\"u}r.
\newblock 2019.
\newblock Multiwoz 2.1: Multi-domain dialogue state corrections and state
  tracking baselines.
\newblock {\em ArXiv}, abs/1907.01669.

\bibitem[\protect\citename{Gao \bgroup et al.\egroup }2019]{Gao2019DialogST}
Shuyang Gao, Abhishek Sethi, Sanchit Agarwal, Tagyoung Chung, and Dilek~Z.
  Hakkani-T{\"u}r.
\newblock 2019.
\newblock Dialog state tracking: A neural reading comprehension approach.
\newblock {\em ArXiv}, abs/1908.01946.

\bibitem[\protect\citename{Goel \bgroup et al.\egroup }2019]{Goel2019HySTAH}
Rahul Goel, Shachi Paul, and Dilek~Z. Hakkani-T{\"u}r.
\newblock 2019.
\newblock Hyst: A hybrid approach for flexible and accurate dialogue state
  tracking.
\newblock {\em ArXiv}, abs/1907.00883.

\bibitem[\protect\citename{Gulcehre \bgroup et al.\egroup
  }2016]{Gulcehre2016Pointing}
Caglar Gulcehre, Sungjin Ahn, Ramesh Nallapati, Bowen Zhou, and Yoshua Bengio.
\newblock 2016.
\newblock Pointing the unknown words.
\newblock In {\em Proceedings of the 54th Annual Meeting of the Association for
  Computational Linguistics (Volume 1: Long Papers)}.

\bibitem[\protect\citename{Hashimoto \bgroup et al.\egroup
  }2016]{Hashimoto2016AJM}
Kazuma Hashimoto, Caiming Xiong, Yoshimasa Tsuruoka, and Richard Socher.
\newblock 2016.
\newblock A joint many-task model: Growing a neural network for multiple nlp
  tasks.
\newblock In {\em EMNLP}.

\bibitem[\protect\citename{Hinton \bgroup et al.\egroup
  }2012]{hinton2012improving}
Geoffrey~E Hinton, Nitish Srivastava, Alex Krizhevsky, Ilya Sutskever, and
  Ruslan~R Salakhutdinov.
\newblock 2012.
\newblock Improving neural networks by preventing co-adaptation of feature
  detectors.
\newblock {\em arXiv preprint arXiv:1207.0580}.

\bibitem[\protect\citename{Hochreiter and Schmidhuber}1997]{hochreiter1997long}
Sepp Hochreiter and J{\"u}rgen Schmidhuber.
\newblock 1997.
\newblock Long short-term memory.
\newblock {\em Neural computation}, 9(8):1735--1780.

\bibitem[\protect\citename{Hosseini-Asl \bgroup et al.\egroup
  }2020]{hosseini2020simple}
Ehsan Hosseini-Asl, Bryan McCann, Chien-Sheng Wu, Semih Yavuz, and Richard
  Socher.
\newblock 2020.
\newblock A simple language model for task-oriented dialogue.
\newblock {\em arXiv preprint arXiv:2005.00796}.

\bibitem[\protect\citename{Kim \bgroup et al.\egroup }2020]{kim2020somdst}
Sungdong Kim, Sohee Yang, Gyuwan Kim, and Sang-woo Lee.
\newblock 2020.
\newblock Efficient dialogue state tracking by selectively overwriting memory.
\newblock In {\em ACL}.

\bibitem[\protect\citename{Kingma and Ba}2014]{kingma2014adam}
Diederik~P Kingma and Jimmy Ba.
\newblock 2014.
\newblock Adam: A method for stochastic optimization.
\newblock {\em arXiv preprint arXiv:1412.6980}.

\bibitem[\protect\citename{Lee \bgroup et al.\egroup }2019]{Lee2019SUMBTSM}
Hwaran Lee, Jinsik Lee, and Tae-Yoon Kim.
\newblock 2019.
\newblock Sumbt: Slot-utterance matching for universal and scalable belief
  tracking.
\newblock In {\em ACL}.

\bibitem[\protect\citename{Luong \bgroup et al.\egroup
  }2015]{luong2015effective}
Minh-Thang Luong, Hieu Pham, and Christopher~D Manning.
\newblock 2015.
\newblock Effective approaches to attention-based neural machine translation.
\newblock {\em arXiv preprint arXiv:1508.04025}.

\bibitem[\protect\citename{Mrksic \bgroup et al.\egroup
  }2016]{Mrksic2016NeuralBT}
Nikola Mrksic, Diarmuid~{\'O} S{\'e}aghdha, Tsung-Hsien Wen, Blaise Thomson,
  and Steve~J. Young.
\newblock 2016.
\newblock Neural belief tracker: Data-driven dialogue state tracking.
\newblock In {\em ACL}.

\bibitem[\protect\citename{Pennington \bgroup et al.\egroup
  }2014]{Pennington2014GloveGV}
Jeffrey Pennington, Richard Socher, and Christopher~D. Manning.
\newblock 2014.
\newblock Glove: Global vectors for word representation.
\newblock In {\em EMNLP}.

\bibitem[\protect\citename{Ramadan \bgroup et al.\egroup
  }2018]{Ramadan2018LargeScaleMB}
Osman Ramadan, Pawel Budzianowski, and Milica Gai.
\newblock 2018.
\newblock Large-scale multi-domain belief tracking with knowledge sharing.
\newblock In {\em ACL}.

\bibitem[\protect\citename{Ren \bgroup et al.\egroup }2019]{Ren2019ScalableAA}
Liliang Ren, Jianmo Ni, and Julian~J. McAuley.
\newblock 2019.
\newblock Scalable and accurate dialogue state tracking via hierarchical
  sequence generation.
\newblock In {\em EMNLP/IJCNLP}.

\bibitem[\protect\citename{See \bgroup et al.\egroup }2017]{see2017get}
Abigail See, Peter~J Liu, and Christopher~D Manning.
\newblock 2017.
\newblock Get to the point: Summarization with pointer-generator networks.
\newblock {\em arXiv preprint arXiv:1704.04368}.

\bibitem[\protect\citename{Serban \bgroup et al.\egroup
  }2015]{Serban2015BuildingED}
Iulian Serban, Alessandro Sordoni, Yoshua Bengio, Aaron~C. Courville, and
  Joelle Pineau.
\newblock 2015.
\newblock Building end-to-end dialogue systems using generative hierarchical
  neural network models.
\newblock In {\em AAAI}.

\bibitem[\protect\citename{Serban \bgroup et al.\egroup }2016]{Serban2016AHL}
Iulian Serban, Alessandro Sordoni, Ryan Lowe, Laurent Charlin, Joelle Pineau,
  Aaron~C. Courville, and Yoshua Bengio.
\newblock 2016.
\newblock A hierarchical latent variable encoder-decoder model for generating
  dialogues.
\newblock In {\em AAAI}.

\bibitem[\protect\citename{Sutskever \bgroup et al.\egroup
  }2014]{Sutskever2014}
Ilya Sutskever, Oriol Vinyals, and Quoc~V Le.
\newblock 2014.
\newblock {Sequence to sequence learning with neural networks}.
\newblock {\em Advances in Neural Information Processing Systems (NIPS)}, pages
  3104--3112.

\bibitem[\protect\citename{Vinyals \bgroup et al.\egroup
  }2015]{Vinyals2015Pointer}
Oriol Vinyals, Meire Fortunato, and Navdeep Jaitly.
\newblock 2015.
\newblock Pointer networks.
\newblock In {\em International Conference on Neural Information Processing
  Systems}.

\bibitem[\protect\citename{Wu \bgroup et al.\egroup
  }2019]{Wu2019TransferableMS}
Chien-Sheng Wu, Andrea Madotto, Ehsan Hosseini-Asl, Caiming Xiong, Richard
  Socher, and Pascale Fung.
\newblock 2019.
\newblock Transferable multi-domain state generator for task-oriented dialogue
  systems.
\newblock In {\em ACL}.

\bibitem[\protect\citename{Xu and Hu}2018]{Xu2018AnEA}
Puyang Xu and Qi~Hu.
\newblock 2018.
\newblock An end-to-end approach for handling unknown slot values in dialogue
  state tracking.
\newblock In {\em ACL}.

\bibitem[\protect\citename{Yang \bgroup et al.\egroup
  }2016]{yang2016hierarchical}
Zichao Yang, Diyi Yang, Chris Dyer, Xiaodong He, Alex Smola, and Eduard Hovy.
\newblock 2016.
\newblock Hierarchical attention networks for document classification.
\newblock In {\em Proceedings of the 2016 Conference of the North American
  Chapter of the Association for Computational Linguistics: Human Language
  Technologies}, pages 1480--1489.

\bibitem[\protect\citename{Zhong \bgroup et al.\egroup
  }2018]{Zhong2018GlobalLocallySD}
Victor Zhong, Caiming Xiong, and Richard Socher.
\newblock 2018.
\newblock Global-locally self-attentive dialogue state tracker.
\newblock {\em ArXiv}, abs/1805.09655.

\end{thebibliography}

\end{document}